\documentclass[conference]{IEEEtran}
\ifdefined\pdfminorversion
\fi
\IEEEoverridecommandlockouts
\usepackage{times}
\usepackage{amsmath}
\usepackage{graphicx}
\usepackage{capt-of}
\usepackage[numbers]{natbib}
\PassOptionsToPackage{hyphens}{url}
\usepackage[bookmarks=true]{hyperref}

\hypersetup{hidelinks,pdftitle={DiMuRe: Diagnosis-Guided Multilevel Recovery for Robotic Task Execution}}

\begin{document}
\title{DiMuRe: Diagnosis-Guided Multilevel Recovery for Robotic Task Execution}
\author{%
\IEEEauthorblockN{Renkai Yi$^{1}$, Yuwei Wu$^{1,2}$, Haoyu Zhang$^{1,2}$, Penghao Dai$^{1}$,\\
Zhi Gao$^{1,2}$, Jin Chen$^{3}$, and Che Sun$^{2}$}
\IEEEauthorblockA{$^{1}$Beijing Institute of Technology, Beijing, China\\
$^{2}$Shenzhen MSU-BIT University, Shenzhen, China\\
$^{3}$AInnovation Technology Group Co., Ltd.\\
\texttt{\{renkaiyi, wuyuwei, zhanghaoyu, penghaodai, gaozhibit\}@bit.edu.cn}\\
\texttt{chen\_jin@bit.edu.cn}; \texttt{sunche@smbu.edu.cn}}
}
\IEEEaftertitletext{%
\begin{minipage}{\textwidth}
\centering
\includegraphics[width=\textwidth]{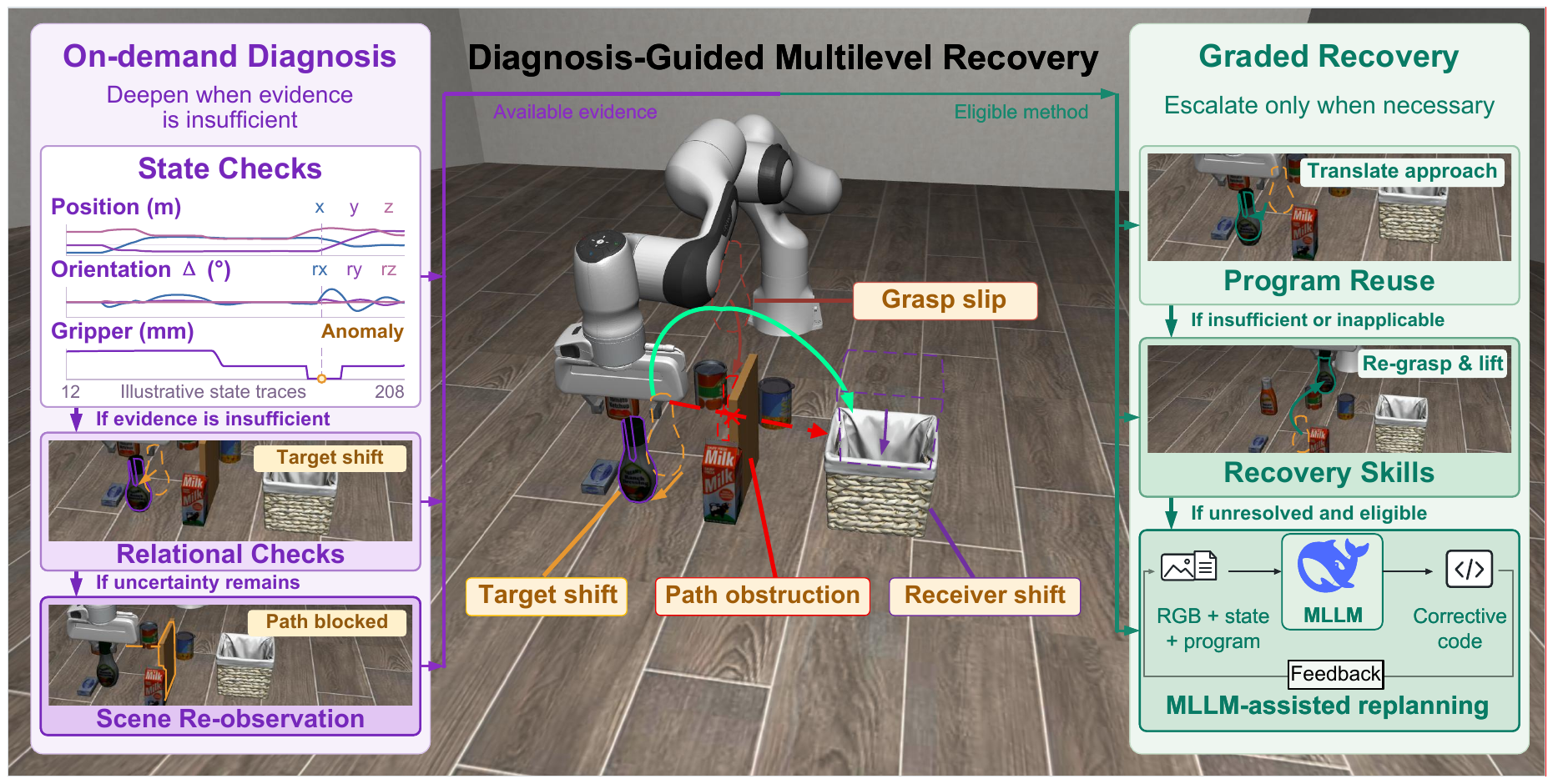}
\captionof{figure}{Conceptual overview of diagnosis-guided recovery in DiMuRe.
The center illustrates target shift, grasp slip, path obstruction, and receiver
shift. On the left, diagnosis starts with state checks and adds relational
checks or new scene observations when more evidence is needed.
On the right, diagnostic evidence and method preconditions determine whether
to reuse the program, invoke a recovery skill, or use MLLM-assisted replanning.
The local recovery examples show a translated approach and a regrasp followed
by a lift. The replanning loop generates corrective code from RGB observations,
execution state, program context, and execution feedback.}
\label{fig:dimure-overview}
\end{minipage}
\vspace{0.7\baselineskip}
}
\maketitle

\begin{abstract}
Disturbances can invalidate the conditions required by a robot's next action
even when much of its task program remains useful. A recovery system must
therefore choose a response suited to the current state.
We present DiMuRe (diagnosis-guided multilevel recovery), a framework for
robot-use agents that uses phase-specific evidence and method preconditions
to select program reuse or predefined recovery skills, including bounded
vision-language-action (VLA) grasping. After the original program exits, a
multimodal large language model (MLLM) can generate corrective code from images
and execution feedback. Local verification checks the outcome of each recovery
attempt. We also introduce LIBERO-Disturb, a benchmark for robustness and
recovery under mid-execution disturbances, supporting program-based agents
and VLA policies. On 801 scheduled disturbed Encore development cases, Local
DiMuRe (program reuse and recovery skills) achieves 77.65\% complete-task
success, compared with 58.68\% for the original program under the same
physical-step allowance. Full DiMuRe achieves 81.90\% with terminal replanning,
an additional step allowance, and early handoff. Local DiMuRe also uses fewer
physical steps per scheduled episode than the original program.
\end{abstract}

\section{Introduction}
When a target object moves during manipulation, a previously valid grasp
command may fail even though much of the task program remains useful.
Robot-use agents express perception, motion, and feedback through executable
programs~\cite{liang2023code,singh2023progprompt}, which can already include
checks and retries. Recovery therefore requires determining which parts of
the existing behavior still apply. If the object has only shifted in position,
translating a recorded grasp segment may suffice. If the grasp assumptions
no longer hold, a different recovery procedure may be needed.

Each recovery option has limits. Program reuse preserves task-specific
motions but depends on the available segments and how they can be adapted.
Predefined recovery skills, including bounded vision-language-action (VLA)
grasping, handle particular failures when their preconditions are met.
A multimodal large language model (MLLM) can compose new action sequences
from visual observations and execution feedback, at the cost of additional
inference and verification. Choosing among these options requires more than
a failure alarm. The system needs evidence that a response is applicable
and must then check whether it restores the state needed to continue the task.
Obtaining this evidence also takes time, so the decision must account for
the cost of further sensing and reasoning.

To study these choices, an evaluation must introduce controlled disturbances
during execution. Specifying a disturbance does not guarantee that it was
applied, had its intended physical effect, or prevented task completion.
Recording these outcomes separately makes it possible to assess both the
original program's resilience and the benefit of added recovery.

To address these challenges, we make two contributions:
\begin{itemize}
\item \textbf{DiMuRe} uses phase-specific diagnostic evidence to select among
program reuse, predefined recovery skills, and MLLM-assisted replanning.
It checks each method's preconditions and verifies the result of recovery
before deciding how to continue.
\item \textbf{LIBERO-Disturb} is a benchmark for mid-execution disturbance
robustness and recovery in robotic manipulation. It separates disturbance
injection and task evaluation from the tested system, supporting both
program-based robot-use agents and VLA policies.
\end{itemize}
The central hypothesis is that different recovery capabilities handle
different failure conditions. The evaluation tests whether combining them
improves task success while preserving the original program's successful
outcomes. It also measures execution time and model use.
Figure~\ref{fig:dimure-overview} illustrates the approach.

\section{Related Work}
\textit{Benchmarks as controlled evaluation settings.}
LIBERO~\cite{liu2023libero} organizes manipulation tasks to study knowledge
transfer in robot learning. LIBERO-Disturb builds on these tasks and their
goal predicates to evaluate responses to disturbances during manipulation.
The disturbance protocol is separate from the tested controller, allowing
evaluation of both program-based agents and VLA policies. DiMuRe is a recovery
framework evaluated within this setting.

\textit{Program structure as a source of recovery capability.}
Program-based control combines perception and action with feedback loops
and precondition checks~\cite{liang2023code,singh2023progprompt}.
These structures provide behaviors that can be reused when execution
deviates from its nominal path. Encore constructs task programs before
evaluation~\cite{kang2026encore}, while interactive coding interfaces
allow observations and execution feedback to inform further
actions~\cite{chu2026openrua,si2026pyrua}. DiMuRe retains existing routines
and uses execution evidence to decide when to invoke additional recovery.

\textit{From execution evidence to recovery decisions.}
Monitoring can assess explicit task constraints~\cite{zhou2025monitor} or
use visual, temporal, and multisensor evidence to detect execution
failures~\cite{duan2025aha,agia2025sentinel,periasami2026frames}. Such evidence
supports both failure explanation and intervention: summarized execution
histories can guide corrective planning~\cite{liu2023reflect}, while policy
uncertainty or grasp-state checks can trigger recovery
actions~\cite{hung2021introspective,wang2026physicalagentic}. The time and
inference required for a decision also matter. AESOP~\cite{sinha2024realtime} connects
fast anomaly detection to slower fallback selection while accounting for
reasoning latency through model predictive control. DiMuRe uses diagnostic evidence to
assess a candidate method's preconditions and verifies whether its execution
restores the state required to continue the task.

\textit{Restoring execution conditions and revising plans.}
Recovery methods can restore a required manipulation state or revise the
actions used to complete a task. Geometric constraints and learned recovery
skills support corrective actions~\cite{huang2025rekep,vats2023efficient,zhao2026flare},
while corrected strategies and recovery experience can be retained for later
reuse~\cite{luo2025roboreflect,liu2026recova}. At the planning level, execution
feedback can support new task plans, motion corrections, or code
repair~\cite{huang2023inner,chen2025promptrecovery,liu2025hycodepolicy}.
The timing of this reasoning also differs: DoReMi~\cite{guo2024doremi}
monitors plan-execution constraints to trigger recovery, whereas
AgentChord~\cite{xu2026agentchord} prepares recovery branches before execution
for activation by compiled monitors. DiMuRe selects among recorded program
segments, predefined skills including VLA grasping, and MLLM-assisted replanning
using method preconditions and local verification. Its evaluation compares
the individual recovery levels with their combined use, reporting task
completion and execution costs under each configuration's stated budget.

\section{Problem Setting}
The agent executes an original task program $\mathsf{P}_{\mathrm{orig}}$.
Its source remains fixed, including its original feedback and retry logic.
Disturbances may displace a target, break a grasp, or move a receiving region.
The task goal remains unchanged, but subsequent commands may fail unless the
affected conditions are restored. Additional recovery is needed when the
original program cannot handle these changes.

Given the task instruction, program context, sensor observations, and action
feedback, the problem is to determine when recovery is needed, which
available method applies, and whether its outcome allows the task to continue.
The objective is to complete the original task within a declared total
physical-step budget $B$, reusing existing behavior where possible and choosing
additional recovery from the available evidence and resources. Online decisions
do not use hidden simulator state, disturbance labels, or the native success
predicate. The native evaluator assesses complete-task success independently.

\section{Layered Diagnosis and Multilevel Recovery}
\begin{figure*}[!t]
\centering
\includegraphics[width=\textwidth]{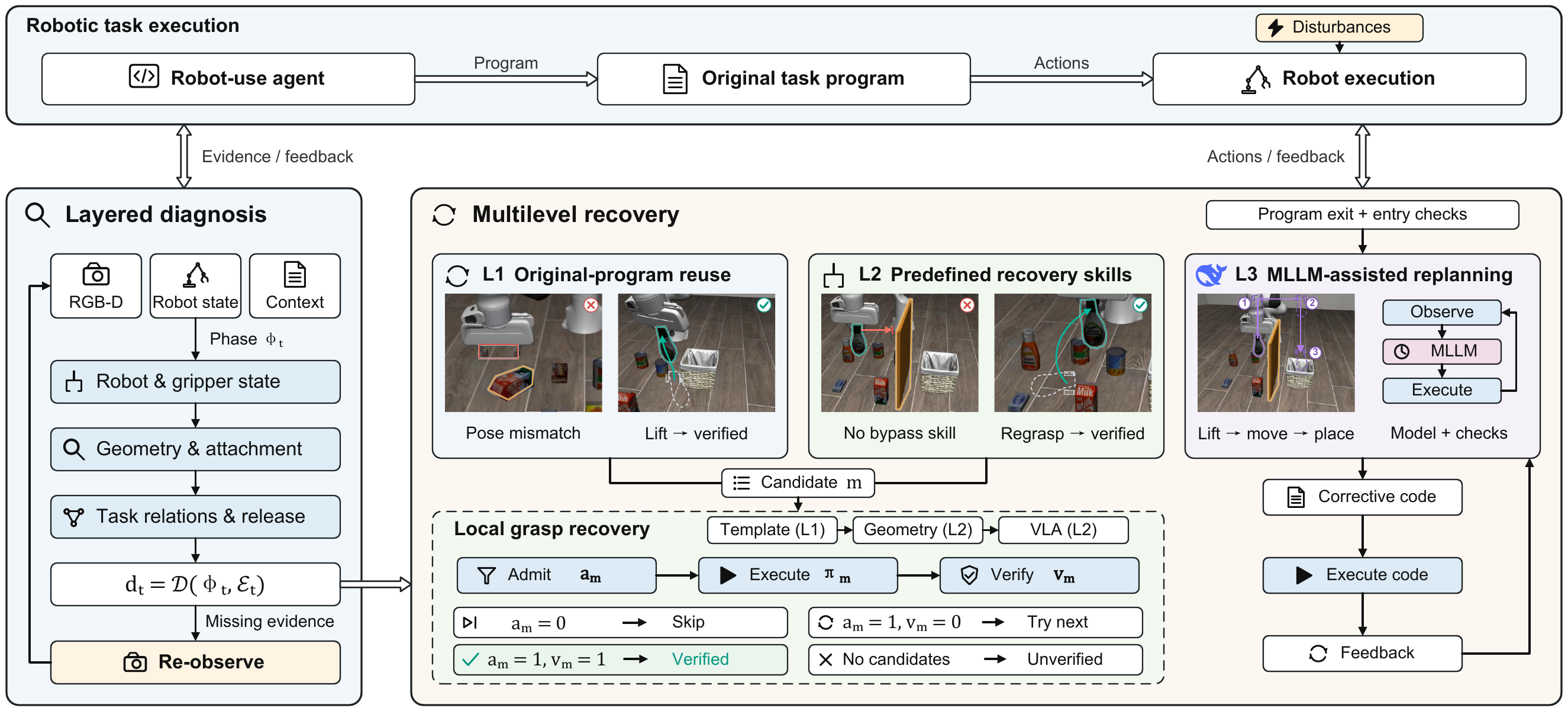}
\caption{DiMuRe system architecture and recovery flow. Phase-specific
diagnosis uses RGB-D observations, robot state, and program context, and
requests new observations when evidence is insufficient. The dashed box
shows local grasp recovery. The system considers template replay (L1),
geometry-based regrasp (L2), and bounded VLA grasping (L2) in that order,
checking admission ($a_m$), executing the method ($\pi_m$), and verifying
the grasp ($v_m$). It skips ineligible candidates and tries the next candidate
if an attempt does not produce a verified grasp. A verified grasp ends the
local search. MLLM-assisted replanning (L3) starts separately after the
original program exits and its entry checks pass. The planner revises
corrective code using new observations and execution feedback.}
\label{fig:dimure-method}
\end{figure*}

Figure~\ref{fig:dimure-method} summarizes how phase-specific diagnosis connects
original-program execution with local recovery and terminal MLLM-assisted
replanning.

\subsection{Program Interface and Observations}
An execution monitor intercepts motion and gripper calls, records grasp and
placement segments, and matches objects and receiving regions named in the
program to sensor observations. In the Encore integration~\cite{kang2026encore},
the monitor uses RGB-D geometry, robot state, and gripper state to track
object identity and motion and to assess grasping and release. It evaluates
relations expressed in the task instruction using measured geometry.
The index $t$ denotes an observation or decision event, which need not advance
physical execution. Positions are expressed in the robot base frame, and
all lengths are in meters. A hat denotes an estimate obtained from RGB-D geometry.

\subsection{Layered Diagnosis: When to Intervene}
The diagnoser checks execution conditions for the current manipulation phase
using robot and gripper state, measured object geometry, and task relations.
It requests additional observations when missing geometry or poor visibility
prevents a decision. The diagnostic record is
\begin{equation}
 d_t=\mathcal D(\phi_t,\mathcal E_t).
\label{eq:phase-diagnosis}
\end{equation}
Here $\mathcal D$ is the rule-based diagnoser,
$\phi_t\in\{\text{grasp},\text{transport},\text{place},\text{verify}\}$ is
the current phase, and $\mathcal E_t$ contains robot state, measured object
geometry, task relations, action feedback, prior verification results, and
budget information. The output $d_t$ contains a diagnostic code,
rule-based confidence, a suggested response, and supporting evidence.
Missing measurements are treated as unknown. The confidence is not a
calibrated probability of success.

During original-program execution, the monitor first checks the gripper
opening, then compares object and tool motion to assess whether the object
is attached to the gripper. Their displacement residual is
\begin{equation}
 e_t^{\mathrm{follow}}=
 \left\|(\hat{\mathbf c}_t-\hat{\mathbf c}_{\mathrm{ref}})
       -(\mathbf p_t-\mathbf p_{\mathrm{ref}})\right\|_2.
\label{eq:hold-consistency}
\end{equation}
The vectors $\hat{\mathbf c}_t$ and $\mathbf p_t$ are the current estimated
object center and measured end-effector position. Subscript $\mathrm{ref}$
denotes the reference measurements stored for the current grasp check.
The Euclidean norm $\|\cdot\|_2$ measures the difference between the object
and tool displacements.
The test requires sufficient visible geometry and tool displacement
$\|\mathbf p_t-\mathbf p_{\mathrm{ref}}\|_2>\tau_{\mathrm{move}}$,
where $\tau_{\mathrm{move}}$ is the minimum tool displacement required
to evaluate attachment. A verified grasp additionally requires
$e_t^{\mathrm{follow}}<\tau_{\mathrm{follow}}$ and
$g_{\min}<g_t<g_{\max}$, where $\tau_{\mathrm{follow}}$ is the
maximum accepted displacement residual, $g_t$ is the measured gripper
opening, and $g_{\min}$ and $g_{\max}$ are its lower and upper admissible
bounds. These thresholds and measurements are lengths in meters.
The monitor also requires evidence that the object is no longer supported.
Insufficient motion or visibility leaves attachment unconfirmed. If the
measurements suggest contact near the receiving region, the diagnoser requests
release verification before classifying the object as held or slipped.
After release, placement checks assess whether the object has detached from
the gripper and is supported by the receiving region.

\subsection{Three Levels of Recovery Capability}
\paragraph{Level 1: Original-program reuse}
Level~1 reuses the original program's feedback, retries, and recorded grasp
and placement segments. When reference geometry is available, grasp replay
translates the recorded waypoints as
\begin{equation}
 \begin{aligned}
 \boldsymbol\delta_t & =
 \begin{bmatrix}
  \hat c_{t,x}-\hat c^{\mathrm{tpl}}_x\\
  \hat c_{t,y}-\hat c^{\mathrm{tpl}}_y\\
  s_t-s^{\mathrm{tpl}}
 \end{bmatrix},\\
 \mathbf q'_j&=\mathbf q_j+\boldsymbol\delta_t,\qquad R'_j=R_j.
 \end{aligned}
\label{eq:template-retargeting}
\end{equation}
Here $j$ indexes motion waypoints in the recorded segment; $\mathbf q_j$ is
the original commanded end-effector position, and the prime denotes its
adapted value. The vector $\boldsymbol\delta_t$ is the measured translation.
The scalars $\hat c_{t,x}$ and $\hat c_{t,y}$ are the horizontal components
of the current object-center estimate. Superscript $\mathrm{tpl}$ denotes
the template's stored object center and support height, while $s_t$ is the
current support height. Any recorded orientation $R_j$, a $3\times3$ rotation
matrix from tool to base coordinates, is preserved. Replay does not add an
orientation command where the recorded segment has none.

Replay requires a recorded segment and an object-center reference. When
reference shape data are available, the system checks shape compatibility
and the change in support height before execution. Otherwise, it omits the
shape comparison and sets the vertical component of $\boldsymbol\delta_t$
to zero. Replay is rejected if the horizontal displacement exceeds its
threshold. Equation~\ref{eq:template-retargeting} describes the waypoint
translation. Execution also includes clearance moves, limits on command
duration, and grasp verification. Updated receiving-region estimates can
similarly be used to adjust placement targets.

\paragraph{Level 2: Predefined recovery skills}
Level~2 uses geometric procedures or learned policies for targeted corrections.
Geometry-based regrasp uses the measured target shape to choose the approach,
gripper orientation, closure, and lift, subject to the skill's object-size limits.
Bounded VLA grasping uses GraspVLA~\cite{deng2025graspvla} as a predefined skill.
The policy receives a grasp instruction and a target hint derived from sensor
observations, and the execution monitor verifies the resulting grasp.
Placement recovery uses an updated receiving-region estimate together with
grasp and release checks. It may reuse level-1 motion segments or invoke
level-2 grasping skills. These skills apply only when their preconditions
hold, and their success still requires verification.

\paragraph{Level 3: MLLM-assisted replanning}
After the original program exits, an MLLM generates code that composes
corrective actions. This code can combine original functions with level-2
skills, including bounded VLA grasping, to attempt recovery beyond a recorded
segment or a single skill. It also adds inference and verification costs.
All three levels use the same robot interface and physical-step accounting.

The source of a recovery behavior is distinct from its physical effect.
For example, replaying an original-program regrasp still counts as a physical
intervention. Verifying a grasp or release does not establish complete-task success.

\subsection{Conditional Switching and Verification}
Local grasp recovery considers candidates in a fixed order. The diagnostic
record $d_t$, manipulation phase $\phi_t$, and method preconditions determine
which candidates can run. A candidate method $m$ is represented by
\begin{equation}
 \mathcal R_m=(\mathcal P_m,\pi_m,\mathcal V_m),
\label{eq:recovery-method}
\end{equation}
where $\mathcal R_m$ denotes the method, $\mathcal P_m$ is its pre-execution
admission predicate, $\pi_m$ its executable recovery routine, and
$\mathcal V_m$ its local grasp-verification rule. The admission predicate
checks that the method and the required geometry or recorded program state
are available, and that any minimum budget requirement is met.
The index $m$ ranges over template replay,
geometry-based regrasp, and bounded VLA grasping, considered in that order.
Template replay belongs to level~1; geometry-based regrasp and VLA grasping
are level-2 recovery skills. The level-3 MLLM planner starts separately
after the original program exits.

The system obtains a fresh target observation before considering each candidate.
If the required geometry is missing, the robot retreats upward and obtains
a refined observation. The procedure ends if the geometry remains unavailable.
Let $a_m=1$ when $\mathcal P_m$ admits execution and $a_m=0$ otherwise.
For an admitted candidate that returns normally, $v_m=1$ only if the execution
monitor verifies the resulting grasp. Missing or unsuccessful verification
gives $v_m=0$. The transition is
\begin{equation}
\operatorname{next}(m)=
\begin{cases}
 \text{skip candidate}, & a_m=0,\\
 \text{return verified grasp}, & a_m=1,\ v_m=1,\\
 \text{advance candidate}, & a_m=1,\ v_m=0.
\end{cases}
\label{eq:recovery-routing}
\end{equation}
Here $\operatorname{next}$ is the decision about how to continue. Both skipping
and advancing move to the next candidate, but only advancing follows execution
of a recovery method. If no candidate remains, the procedure returns without
a verified grasp. Native termination or budget exhaustion stops the procedure.
Candidate attempts share a task-wide integer limit $A_{\max}$. Each candidate
consumes one attempt before its preconditions are checked, even if it is then
skipped. No further candidates are considered once the limit is reached.
After a grasp is verified, movement toward the adapted target is checked
separately. Equation~\ref{eq:recovery-routing} describes grasp recovery.
Placement recovery uses its own release checks.

\paragraph{Shared budget and completion reserve}
Let $n_t$ be the cumulative number of physical control steps consumed by
event $t$, and let $B_t\leq B$ be the active ceiling. Here $B$ is the total
budget declared for the configuration. The remaining budget is
$b_t=\max(0,B_t-n_t)$. The evaluation protocol may assign different ceilings
to the native and terminal phases.
Within a recovery procedure, ordinary motion, gripper, and settling commands
are admitted only if $n_t+\bar c(u)\leq C$. The ceiling is fixed on entry as
\begin{equation}
 C=\min\bigl\{B_t-r,\ n_{\mathrm{start}}+C_{\max},\ C_{\mathrm{outer}}\bigr\}.
\label{eq:recovery-budget}
\end{equation}
During the procedure, the cumulative physical-step count may not exceed $C$.
The nonnegative integer $r$ reserves steps for task completion,
$n_{\mathrm{start}}$ is the count at procedure entry, and $C_{\max}$ caps
the additional steps the procedure may consume. An enclosing procedure
imposes $C_{\mathrm{outer}}$, set to $+\infty$ if no such procedure exists.
The command $u$ specifies a primitive and its requested duration.
The implementation estimates its maximum step cost as $\bar c(u)$.
All budget quantities are measured in physical steps. Commands are also
rejected when native execution has stopped or $n_t\geq B_t$.

Procedure entry additionally requires $b_t\geq b_{\min}$, where $b_{\min}$
is that procedure's minimum remaining physical-step budget. This threshold,
the reserve $r$, and the cap $C_{\max}$ are specified for each procedure.
Nested VLA calls are also subject to tool-specific entry and completion-reserve
checks. Recovery parameters and invocation limits are provided in
Appendix~\ref{app:implementation-parameters}.

\subsection{Terminal MLLM-Assisted Replanning}
Planner entry and round continuation require
$b_t\geq b_{\mathrm{entry}}^{\mathrm{plan}}$ and
$b_t\geq b_{\mathrm{round}}^{\mathrm{plan}}$, respectively, where the two
thresholds specify the minimum remaining physical-step budgets for these
decisions and $\mathrm{plan}$ identifies the terminal planner.
At each round, the MLLM generates corrective code from current RGB images,
structured state evidence, program context, and execution history. After the
code runs, the planner receives updated observations, local verification
results, and action feedback to revise its next actions. If the model proposes
stopping without geometric evidence of completion, the system requests another
review, subject to the review limit.

\section{LIBERO-Disturb: Mid-Execution Disturbance Benchmark}
\label{sec:libero-disturb}

\begin{figure*}[!t]
\centering
\includegraphics[width=\textwidth]{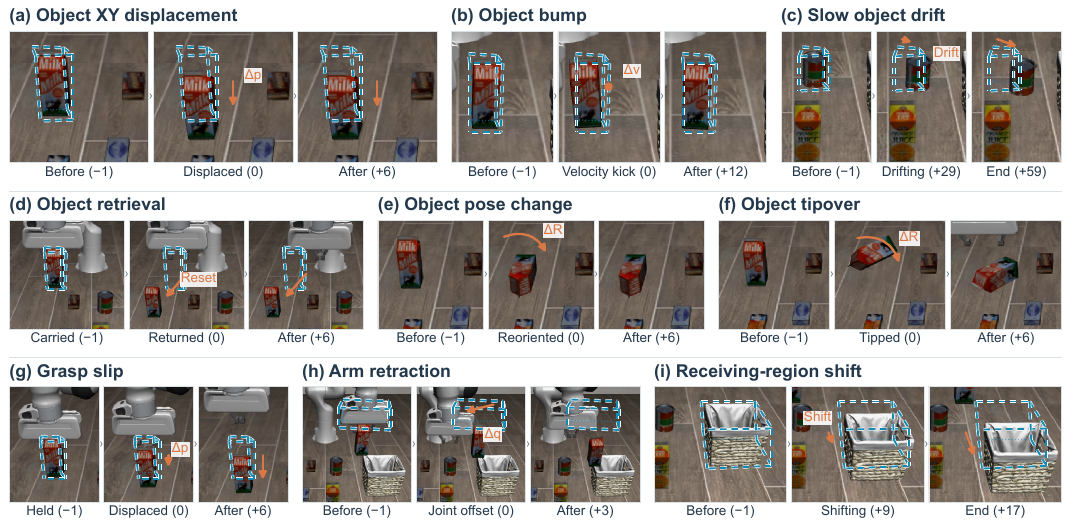}
\caption{Nine mid-execution disturbances in LIBERO-Disturb, shown in recorded
Encore episodes: (a)--(d) object location and motion, (e)--(f) object pose,
(g)--(h) grasp and robot state, and (i) receiving-region location.
The three images in each example use the same camera and crop. Blue dashed
wireframes mark the approximate pre-intervention position. Orange annotations indicate
position ($\Delta p$), velocity ($\Delta v$), orientation ($\Delta R$), or
joint-position ($\Delta q$) changes and their visible effects; grey chevrons
indicate time order. Values in parentheses are relative native control-step
indices $\Delta k$. Zero marks the first recorded post-injection frame, captured
after one native control step following the first intervention. The overlays
are qualitative visual guides.}
\label{fig:libero-disturb-nine}
\end{figure*}

\paragraph{Benchmark overview}
LIBERO-Disturb evaluates whether a robot can complete a manipulation task
when a disturbance changes its execution conditions. Built on LIBERO tasks and goal
predicates~\cite{liu2023libero}, it introduces controlled disturbances while
retaining the original instruction and task objective. This setting tests
both the resilience of existing behavior and the benefit of added recovery.
Disturbance injection and task evaluation are separated from the tested
controller, supporting program-based agents and VLA policies through their
respective native execution interfaces.

\paragraph{Disturbance design}
The nine disturbance types affect four aspects of execution
(Fig.~\ref{fig:libero-disturb-nine}). Four change object location or motion:
\emph{XY displacement} changes position, \emph{object bump} adds a velocity
increment, \emph{slow drift} applies repeated small translations, and
\emph{object retrieval} returns a carried object to its saved initial
position.

Two disturbances affect pose: \emph{pose change} applies a prescribed resting
orientation, and \emph{tipover} imposes a tipping rotation. These changes
challenge the geometry assumed during approach and grasping. \emph{Grasp slip}
displaces a held object relative to the gripper, while \emph{arm retraction}
perturbs arm joint positions, challenging attachment and motion continuity.
Finally, \emph{receiving-region shift} moves the destination during transport,
which can invalidate a previously selected placement target. Affected objects follow
the declared selection rule and need not be the task's named target.
Together, these disturbances test when the original behavior remains useful
and when additional recovery is needed.

\paragraph{Evaluation protocol}
Each case specifies the task initialization, random seeds, disturbance rule,
and execution limits. An intervention is applied only when its trigger and
eligibility conditions are met; one event may extend over several control
steps. The evaluation records disturbance application, its observed physical
effect, and native complete-task success separately. A disturbance with a
verified physical effect need not cause an execution fault. All scheduled
cases remain in the task-success denominator, and unapplied events are
reported separately. Within a fixed system and protocol, configurations are
paired by scheduled case to compare task success, physical steps, elapsed time,
and model use. Existing Encore and VLA results are reported
separately because their native protocols and evaluated coverage differ.
Trigger rules, intervention parameters, and system-specific coverage
are specified in Appendix~\ref{app:benchmark-evidence}.

\section{Experiments}
DiMuRe is evaluated on Encore for task completion, recovery-level ablation,
execution costs, and anomaly response. The separate native-policy
evaluation of $\pi_{0.5}$~\cite{physicalintelligence2025pi05} is reported in
Appendix~\ref{app:benchmark-evidence}.
\subsection{Experimental Setup}
The evaluation compares five configurations on the same 801 scheduled
disturbed cases, using 30 fixed Encore task programs and three reset seeds.
All configurations use the same disturbance injector and native evaluator
and retain the original program's checks and retries.
Original has no added DiMuRe recovery. L1-only adds program reuse, and L2-only
adds predefined geometry and VLA skills. Local (L1+L2) enables both, while
Full (L1+L2+L3) also enables terminal MLLM replanning. L2-only disables the
mixed placement retry because that procedure replays program segments.

Original, L1-only, L2-only, and Local share the same native physical-step
allowance. Full has an additional terminal allowance and can hand off to the
planner early. Its comparison with Local therefore does not isolate the
effect of L3 under an equal budget. Original, Local, and Full also run on
90 clean cases. The disturbed subset follows an earlier development selection.
The full 891-case set and both receiving-region-shift amplitudes are reported in
Table~\ref{tab:supp-encore-sensitivity}. Run settings are given in
Appendix~\ref{app:pilot-protocol}.

\subsection{Evaluation Metrics}
\paragraph{Task performance}
Task success rate (SR) is the fraction of scheduled episodes completed
according to the native evaluator. The denominator includes episodes in
which the disturbance trigger is never reached or the intervention has no
effect. Changes in SR ($\Delta$SR) are measured relative to Original on the
same scheduled cases and reported in percentage points (pp), using unrounded rates.

\paragraph{Execution costs}
Task time includes online sensing, inference, service waits, execution,
and recovery. Physical steps count executed native control steps, including
recovery. All-case means include failed episodes. Successful episodes are
measured up to the first native success, and successful-case means use each
configuration's own successful subset. These subsets can differ.
Wall-clock comparisons are descriptive because configurations shared services
and their runs were not fully interleaved.

\paragraph{Recorded recovery time}
For each episode, recorded local and terminal recovery intervals are summed
separately, counting only time within the task interval. Each mean uses the
episodes that entered the corresponding stage, including failed episodes.
The logs do not isolate diagnosis and adaptations made outside these recorded
intervals. Detailed timing boundaries are given in
Appendix~\ref{sec:supp-recovery-timing}.

\paragraph{Anomaly response and diagnosis}
Anomaly-response coverage is measured on fault episodes that can be labeled
independently. It is the fraction with an explicit alarm or an observation-driven
grasp or placement adjustment within the fault window. Task SR after response
is the fraction of responding episodes that subsequently achieve native task
success. Diagnostic accuracy separately requires the correct fault category
and object role. Detailed labeling rules, diagnostic results, paired outcomes,
and terminal MLLM use are reported in
Appendix~\ref{app:pilot-protocol}.

\subsection{Task Performance and Ablation}

\paragraph{Overall disturbance robustness}
Table~\ref{tab:pilot-results} reports disturbed-task SRs of 58.68\% for
Original, 77.65\% for Local, and 81.90\% for Full. All three configurations
complete all 90 clean cases, so no loss of clean-task success is observed
in these runs.

\begin{table}[t]
\centering
\caption{Native task SR (\%) on 90 clean and 801 scheduled disturbed
Encore cases per configuration.}
\label{tab:pilot-results}
\footnotesize
\setlength{\tabcolsep}{5pt}
\begin{tabular}{@{}lrr@{}}
\hline
Configuration & \shortstack{Clean\\SR (\%)}
 & \shortstack{Disturbed\\SR (\%)} \\
\hline
\multicolumn{3}{@{}l}{Encore} \\
Original & 100.00 & 58.68 \\
Local DiMuRe (L1+L2) & 100.00 & 77.65 \\
Full DiMuRe & 100.00 & 81.90 \\
\hline
\end{tabular}
\end{table}

\begin{table*}[t]
\centering
\caption{Recovery-level ablation, execution costs, and process analysis.
Panel (a) uses 801 scheduled disturbed cases; $\Delta$SR is relative to
Original. All-case means include failures; successful-case means use each
configuration's successful subset. Full has an additional allowance and
distinct handoff.}
\label{tab:encore-ablation}
\small
\setlength{\tabcolsep}{4pt}
\begin{tabular*}{\textwidth}{@{\extracolsep{\fill}}lrrrrrr@{}}
\multicolumn{7}{@{}l}{\textbf{(a) Task outcomes and execution costs}} \\[2pt]
\hline
Configuration & \shortstack{SR\\(\%)}
 & \shortstack{$\Delta$SR\\(pp)}
 & \multicolumn{2}{c}{Mean task time (s)}
 & \multicolumn{2}{c}{Mean physical steps} \\
\cline{4-5}\cline{6-7}
 & & & All cases & Successful cases & All cases & Successful cases \\
\hline
\multicolumn{7}{@{}l}{Encore: shared native allowance} \\
Original program & 58.68 & 0.00 & 4.65 & 2.93 & 328.90 & 213.46 \\
L1-only & 73.66 & $+14.98$ & 5.24 & 3.68 & 317.32 & 229.20 \\
L2-only & 65.54 & $+6.87$ & 5.91 & 3.95 & 349.72 & 229.49 \\
Local DiMuRe (L1+L2) & 77.65 & $+18.98$ & 5.47 & 3.94 & 310.20 & 226.89 \\
\hline
\multicolumn{7}{@{}l}{Full configuration: additional allowance and distinct handoff} \\
Full DiMuRe & 81.90 & $+23.22$ & 19.59 & 7.05 & 361.30 & 248.25 \\
\hline
\end{tabular*}
\par\medskip
\begin{tabular*}{\textwidth}{@{\extracolsep{\fill}}lrrrrrr@{}}
\multicolumn{7}{@{}l}{\textbf{(b) Anomaly response (161/801 cases) and recovery timing (all 801 logs)}} \\[2pt]
\hline
Configuration & \shortstack{Response\\coverage (\%)}
 & \shortstack{Task SR after\\response (\%)}
 & \multicolumn{2}{c}{Local recovery (L1+L2)}
 & \multicolumn{2}{c}{Terminal recovery (L3)} \\
\cline{4-5}\cline{6-7}
 & & & Episodes & Mean (s) & Episodes & Mean (s) \\
\hline
Local DiMuRe (L1+L2) & 126/161 (78.26) & 94/126 (74.60) & 271 & 1.80 & 0 & N/A \\
Full DiMuRe & 126/161 (78.26) & 101/126 (80.16) & 271 & 1.78 & 180 & 62.86 \\
\hline
\end{tabular*}
\par\smallskip
\begin{minipage}{\textwidth}
\footnotesize
Response entries are count/denominator (\%). Recovery time is summed within
each episode up to the task-time endpoint, then averaged over episodes that
entered the stage, including failures. It includes sensing, reasoning, waits,
actions, and verification within the recorded intervals. Monitoring and
adaptation outside those intervals are not measured separately.
Local and L3 episode sets may overlap. N/A denotes no entry.
\end{minipage}
\end{table*}

\paragraph{Recovery-level ablation}
L1-only and L2-only exceed Original by 14.98 and 6.87 percentage points,
respectively (Table~\ref{tab:encore-ablation}(a)). Combining them in Local
gives a larger gain of 18.98 points. L1-only completes more grasp-slip cases,
whereas L2-only completes more object-pose-change cases
(Table~\ref{tab:supp-single-level-types}). These results suggest that the
two recovery levels address different failures.

Across all scheduled cases, Local uses fewer physical steps than Original
on average (310.20 versus 328.90), but takes more time (5.47 versus 4.65\,s).
Among successful episodes, Local averages 3.94\,s and 226.89 steps, compared
with 2.93\,s and 213.46 steps for Original. Thus, Local's lower step count
over all cases does not imply a lower mean among successful episodes,
whose membership differs between configurations.

Full exceeds Local by a further 4.24 percentage points in SR. It also has
the highest all-case means: 19.59\,s and 361.30 steps. Among successful
episodes, the means are 7.05\,s and 248.25 steps. Full therefore completes
more tasks at greater execution cost, under its additional allowance and
early-handoff rules.

\subsection{Anomaly Response and Recovery Time}
Saved ground-truth trajectories and video from the same trial allow fault
windows to be identified in 161 of 801 episodes per configuration.
Episodes without sufficient evidence are left unlabeled, not treated as normal.
In Table~\ref{tab:encore-ablation}(b), response coverage is 78.26\% for both
configurations, while subsequent task SR among responding episodes is
74.60\% for Local and 80.16\% for Full.

Recovery timing uses all 801 episode logs per configuration, regardless of
whether a fault label is available. Each mean covers the episodes that enter
that stage. Because these episode sets differ, the means do not compare
recovery speed on matched inputs. Timing boundaries and coverage are given in
Appendix~\ref{sec:supp-recovery-timing}.

\section{Limitations and Discussion}
Occlusion and incorrect object matching can cause errors in both method
selection and local verification. A fixed candidate order may also exhaust
the budget before a suitable VLA skill or the MLLM planner can be used.
A more reliable task-completion check is needed to decide whether terminal
replanning is necessary. The contributions of its perception, memory, and
reasoning components also require further ablation. Any stored recovery
experience must be fixed before evaluation and kept separate from test cases.

The single-level comparisons do not fully isolate each component because
placement retries can combine program replay with recovery skills. Full's
additional allowance and early handoff also prevent attributing its gains
over Local to L3 alone. The retrospective diagnostic analysis covers only
fault windows that can be labeled independently. Normal states and faults
across the full evaluation set have not been exhaustively annotated.
Further controls should use equal physical-step budgets and identical
diagnostic inputs.

The existing Encore and VLA evaluations use different disturbance versions
and execution protocols. They show that disturbances can be evaluated with
both program-based agents and VLA policies, but direct performance comparisons
require aligned conditions and scoring. Current evidence is limited to simulation.

\section{Conclusion}
LIBERO-Disturb evaluates robustness and recovery under mid-execution
disturbances, recording disturbance application, physical effects, and task
outcomes separately. DiMuRe uses diagnostic evidence to select program reuse,
predefined recovery skills including bounded VLA grasping, or MLLM-assisted
replanning, and verifies the result before continuing. In the Encore
development experiments, Local DiMuRe achieves higher native complete-task
success than the original program under the same step allowance. Full DiMuRe
achieves further gains with a larger allowance and early handoff, at the cost
of additional online time and terminal model use.

\bibliographystyle{mildunsrtnat}
\bibliography{references}

@inproceedings{liang2023code,
  author = {Liang, Jacky and Huang, Wenlong and Xia, Fei and Xu, Peng and
            Hausman, Karol and Ichter, Brian and Florence, Pete and Zeng, Andy},
  title = {\href{https://doi.org/10.1109/ICRA48891.2023.10160591}{Code as Policies: Language model programs for embodied control}},
  booktitle = {IEEE International Conference on Robotics and Automation (ICRA)},
  pages = {9493--9500},
  year = {2023},
  doi = {10.1109/ICRA48891.2023.10160591}
}

@inproceedings{huang2023inner,
  author = {Huang, Wenlong and Xia, Fei and Xiao, Ted and Chan, Harris and
            Liang, Jacky and Florence, Pete and Zeng, Andy and Tompson, Jonathan and
            Mordatch, Igor and Chebotar, Yevgen and Sermanet, Pierre and
            Jackson, Tomas and Brown, Noah and Luu, Linda and Levine, Sergey and
            Hausman, Karol and Ichter, Brian},
  title = {\href{https://proceedings.mlr.press/v205/huang23c.html}{Inner Monologue: Embodied reasoning through planning with language models}},
  booktitle = {Proceedings of the 6th Conference on Robot Learning},
  series = {Proceedings of Machine Learning Research},
  volume = {205},
  pages = {1769--1782},
  year = {2023},
  publisher = {PMLR}
}

@inproceedings{huang2025rekep,
  author = {Huang, Wenlong and Wang, Chen and Li, Yunzhu and Zhang, Ruohan and
            Fei-Fei, Li},
  title = {\href{https://proceedings.mlr.press/v270/huang25g.html}{ReKep: Spatio-temporal reasoning of relational keypoint constraints for robotic manipulation}},
  booktitle = {Proceedings of the 8th Conference on Robot Learning},
  series = {Proceedings of Machine Learning Research},
  volume = {270},
  pages = {4573--4602},
  year = {2025},
  publisher = {PMLR}
}

@inproceedings{agia2025sentinel,
  author = {Agia, Christopher and Sinha, Rohan and Yang, Jingyun and Cao, Ziang and
            Antonova, Rika and Pavone, Marco and Bohg, Jeannette},
  title = {\href{https://proceedings.mlr.press/v270/agia25a.html}{Unpacking failure modes of generative policies: Runtime monitoring of consistency and progress}},
  booktitle = {Proceedings of the 8th Conference on Robot Learning},
  series = {Proceedings of Machine Learning Research},
  volume = {270},
  pages = {689--723},
  year = {2025},
  publisher = {PMLR}
}

@inproceedings{duan2025aha,
  author = {Duan, Jiafei and Pumacay, Wilbert and Kumar, Nishanth and Wang, Yi Ru and
            Tian, Shulin and Yuan, Wentao and Krishna, Ranjay and Fox, Dieter and
            Mandlekar, Ajay and Guo, Yijie},
  title = {\href{https://proceedings.iclr.cc/paper_files/paper/2025/hash/70a06501001e1820fd1eb9ee821302d2-Abstract-Conference.html}{AHA: A vision-language-model for detecting and reasoning over failures in robotic manipulation}},
  booktitle = {International Conference on Learning Representations},
  year = {2025}
}

@inproceedings{zhou2025monitor,
  author = {Zhou, Enshen and Su, Qi and Chi, Cheng and Zhang, Zhizheng and
            Wang, Zhongyuan and Huang, Tiejun and Sheng, Lu and Wang, He},
  title = {\href{https://doi.org/10.1109/CVPR52734.2025.00649}{Code-as-Monitor: Constraint-aware visual programming for reactive and proactive robotic failure detection}},
  booktitle = {Proceedings of the IEEE/CVF Conference on Computer Vision and Pattern Recognition},
  pages = {6919--6929},
  year = {2025},
  doi = {10.1109/CVPR52734.2025.00649}
}

@inproceedings{liu2023reflect,
  author = {Liu, Zeyi and Bahety, Arpit and Song, Shuran},
  title = {\href{https://proceedings.mlr.press/v229/liu23g.html}{REFLECT: Summarizing robot experiences for failure explanation and correction}},
  booktitle = {Proceedings of the 7th Conference on Robot Learning},
  series = {Proceedings of Machine Learning Research},
  volume = {229},
  pages = {3468--3484},
  year = {2023},
  publisher = {PMLR}
}

@inproceedings{hung2021introspective,
  author = {Hung, Chia-Man and Sun, Li and Wu, Yizhe and Havoutis, Ioannis and
            Posner, Ingmar},
  title = {\href{https://doi.org/10.1109/ICRA48506.2021.9561749}{Introspective visuomotor control: Exploiting uncertainty in deep visuomotor control for failure recovery}},
  booktitle = {IEEE International Conference on Robotics and Automation (ICRA)},
  pages = {6293--6299},
  year = {2021},
  doi = {10.1109/ICRA48506.2021.9561749}
}

@inproceedings{vats2023efficient,
  author = {Vats, Shivam and Likhachev, Maxim and Kroemer, Oliver},
  title = {\href{https://doi.org/10.1109/ICRA48891.2023.10160382}{Efficient recovery learning using model predictive meta-reasoning}},
  booktitle = {IEEE International Conference on Robotics and Automation (ICRA)},
  pages = {7258--7264},
  year = {2023},
  doi = {10.1109/ICRA48891.2023.10160382}
}

@inproceedings{sinha2024realtime,
  author = {Sinha, Rohan and Elhafsi, Amine and Agia, Christopher and
            Foutter, Matt and Schmerling, Edward and Pavone, Marco},
  title = {\href{https://www.roboticsproceedings.org/rss20/p114.html}{Real-time anomaly detection and reactive planning with large language models}},
  booktitle = {Proceedings of Robotics: Science and Systems},
  year = {2024},
  doi = {10.15607/RSS.2024.XX.114}
}

@inproceedings{xu2026agentchord,
  author = {Xu, Sheng and Jin, Ruixing and Zhou, Huayi and Yue, Bo and
            Qiao, Guanren and Tai, Yunxin and Deng, Yueci and Jia, Kui and
            Liu, Guiliang},
  title = {\href{https://www.roboticsproceedings.org/rss22/p180.html}{From reaction to anticipation: Proactive failure recovery through agentic task graph for robotic manipulation}},
  booktitle = {Proceedings of Robotics: Science and Systems},
  year = {2026},
  doi = {10.15607/RSS.2026.XXII.180}
}

@inproceedings{zhao2026flare,
  author = {Zhao, Ganlong and Tang, Zijia and Chen, Xingping and Kuang, Zhanghui and
            Tian, Ye and Li, Guanbin},
  title = {\href{https://openaccess.thecvf.com/content/CVPR2026/html/Zhao_FLARE_A_Failure-Aware_Framework_for_Autonomous_Correction_and_Recovery_in_CVPR_2026_paper.html}{FLARE: A failure-aware framework for autonomous correction and recovery in visual-language robotic manipulation}},
  booktitle = {Proceedings of the IEEE/CVF Conference on Computer Vision and Pattern Recognition},
  pages = {22391--22401},
  year = {2026}
}

@inproceedings{singh2023progprompt,
  author = {Singh, Ishika and Blukis, Valts and Mousavian, Arsalan and Goyal, Ankit and
            Xu, Danfei and Tremblay, Jonathan and Fox, Dieter and Thomason, Jesse and
            Garg, Animesh},
  title = {\href{https://doi.org/10.1109/ICRA48891.2023.10161317}{ProgPrompt: Generating situated robot task plans using large language models}},
  booktitle = {IEEE International Conference on Robotics and Automation (ICRA)},
  pages = {11523--11530},
  year = {2023},
  doi = {10.1109/ICRA48891.2023.10161317}
}

@inproceedings{guo2024doremi,
  author = {Guo, Yanjiang and Wang, Yen-Jen and Zha, Lihan and Chen, Jianyu},
  title = {\href{https://doi.org/10.1109/IROS58592.2024.10802284}{DoReMi: Grounding language model by detecting and recovering from plan-execution misalignment}},
  booktitle = {IEEE/RSJ International Conference on Intelligent Robots and Systems (IROS)},
  pages = {12124--12131},
  year = {2024},
  doi = {10.1109/IROS58592.2024.10802284}
}

@misc{kang2026encore,
  author = {Kang, Yifan and Wang, Zihan and Fan, Zhiwen and Liu, Bangya},
  title = {\href{https://arxiv.org/abs/2609.37359}{Encore: Few-shot agentic discovery of manipulation strategies}},
  howpublished = {arXiv preprint arXiv:2609.37359},
  year = {2026},
  doi = {10.48550/arXiv.2609.37359}
}

@misc{chu2026openrua,
  author = {Chu, Zhaoyang and Barr, Earl T. and Le Goues, Claire and O'Hearn, Peter and
            Harman, Mark and Sarro, Federica and Ye, He},
  title = {\href{https://arxiv.org/abs/2610.02459}{OpenRUA: Robot-use agents are zero-shot visuomotor policies}},
  howpublished = {arXiv preprint arXiv:2610.02459},
  year = {2026},
  doi = {10.48550/arXiv.2610.02459}
}

@misc{si2026pyrua,
  author = {Si, Ruiyang and Bi, Jianxin and Yang, Shunyu and Ni, Rui and
            Huang, Wenbo and Wang, Qiang and Jiang, Shulong and Zhang, Yintianrun and
            Ruan, Feixiang and Zhang, Kewei and Wang, Duomin and Li, Xiuyu and
            Feng, Haiwen and Dong, Zhen and Zhou, Daquan},
  title = {\href{https://arxiv.org/abs/2610.01939}{Fewer tokens, better action: GPT-6 Astra robot agents with 14\% higher success rate but 65\% fewer tokens}},
  howpublished = {arXiv preprint arXiv:2610.01939},
  year = {2026},
  doi = {10.48550/arXiv.2610.01939}
}

@misc{liu2025hycodepolicy,
  author = {Liu, Yibin and Liang, Zhixuan and Chen, Zanxin and Chen, Tianxing and
            Hu, Mengkang and Dong, Wanxi and Xu, Congsheng and Han, Zhaoming and
            Qin, Yusen and Mu, Yao},
  title = {\href{https://arxiv.org/abs/2508.02629}{HyCodePolicy: Hybrid language controllers for multimodal monitoring and decision in embodied agents}},
  howpublished = {arXiv preprint arXiv:2508.02629},
  year = {2025},
  doi = {10.48550/arXiv.2508.02629}
}

@misc{wang2026physicalagentic,
  author = {Wang, Wenze and Hosseinzadeh, Mehdi and Dayoub, Feras},
  title = {\href{https://arxiv.org/abs/2604.07395}{A physical agentic loop for language-guided grasping with execution-state monitoring}},
  howpublished = {arXiv preprint arXiv:2604.07395},
  year = {2026},
  doi = {10.48550/arXiv.2604.07395}
}

@misc{luo2025roboreflect,
  author = {Luo, Zhen and Yang, Yixuan and Zhang, Yanfu and Zheng, Feng},
  title = {\href{https://arxiv.org/abs/2501.09307}{RoboReflect: A robotic reflective reasoning framework for grasping ambiguous-condition objects}},
  howpublished = {arXiv preprint arXiv:2501.09307},
  year = {2025},
  doi = {10.48550/arXiv.2501.09307}
}

@misc{liu2026recova,
  author = {Liu, Isabella and Cheng, An-Chieh and Bjorck, Johan and Yu, Zhiding and
            Yin, Hongxu and Kautz, Jan and Fan, Linxi and Zhu, Yuke and Liu, Sifei},
  title = {\href{https://arxiv.org/abs/2610.01178}{Recova: Agent-guided failure recovery for autonomous robotic manipulation}},
  howpublished = {arXiv preprint arXiv:2610.01178},
  year = {2026},
  doi = {10.48550/arXiv.2610.01178}
}

@inproceedings{chen2025promptrecovery,
  author = {Chen, Hongyi and Yao, Yunchao and Liu, Ruixuan and Liu, Changliu and
            Ichnowski, Jeffrey},
  title = {\href{https://doi.org/10.23919/ACC63710.2025.11107751}{Robot failure recovery using vision-language models with optimized prompts}},
  booktitle = {American Control Conference (ACC)},
  pages = {1983--1988},
  year = {2025},
  doi = {10.23919/ACC63710.2025.11107751}
}

@misc{periasami2026frames,
  author = {Periasami, Ajay Vikram and Luo, Xinyuan and Li, Haoyu and Cheng, Xianyi},
  title = {\href{https://arxiv.org/abs/2609.22538}{FRAMES: Failure recovery and monitoring of embodied skills for humanoid loco-manipulation}},
  howpublished = {arXiv preprint arXiv:2609.22538},
  year = {2026},
  doi = {10.48550/arXiv.2609.22538}
}

@article{physicalintelligence2025pi05,
  author = {{Physical Intelligence} and Black, Kevin and Brown, Noah and
            Darpinian, James and Dhabalia, Karan and Driess, Danny and
            Esmail, Adnan and Equi, Michael and Finn, Chelsea and Fusai, Niccolo and
            Galliker, Manuel Y. and Ghosh, Dibya and Groom, Lachy and
            Hausman, Karol and Ichter, Brian and Jakubczak, Szymon and
            Jones, Tim and Ke, Liyiming and LeBlanc, Devin and Levine, Sergey and
            Li-Bell, Adrian and Mothukuri, Mohith and Nair, Suraj and Pertsch, Karl and
            Ren, Allen Z. and Shi, Lucy Xiaoyang and Smith, Laura and
            Springenberg, Jost Tobias and Stachowicz, Kyle and Tanner, James and
            Vuong, Quan and Walke, Homer and Walling, Anna and Wang, Haohuan and
            Yu, Lili and Zhilinsky, Ury},
  title = {\href{https://arxiv.org/abs/2504.16054}{$\pi_{0.5}$: A vision-language-action model with open-world generalization}},
  journal = {arXiv preprint arXiv:2504.16054},
  year = {2025},
  doi = {10.48550/arXiv.2504.16054}
}

@inproceedings{liu2023libero,
  author = {Liu, Bo and Zhu, Yifeng and Gao, Chongkai and Feng, Yihao and
            Liu, Qiang and Zhu, Yuke and Stone, Peter},
  title = {\href{https://proceedings.neurips.cc/paper_files/paper/2023/hash/8c3c666820ea055a77726d66fc7d447f-Abstract-Datasets_and_Benchmarks.html}{LIBERO: Benchmarking knowledge transfer for lifelong robot learning}},
  booktitle = {Advances in Neural Information Processing Systems},
  volume = {36},
  pages = {44776--44791},
  year = {2023},
  doi = {10.52202/075280-1939}
}

@inproceedings{deng2025graspvla,
  author = {Deng, Shengliang and Yan, Mi and Wei, Songlin and Ma, Haixin and
            Yang, Yuxin and Chen, Jiayi and Zhang, Zhiqi and Yang, Taoyu and
            Zhang, Xuheng and Cui, Heming and Zhang, Zhizheng and Wang, He},
  title = {\href{https://proceedings.mlr.press/v305/deng25a.html}{GraspVLA: A grasping foundation model pre-trained on billion-scale synthetic action data}},
  booktitle = {Proceedings of the 9th Conference on Robot Learning},
  series = {Proceedings of Machine Learning Research},
  volume = {305},
  pages = {1004--1029},
  year = {2025},
  publisher = {PMLR}
}

@inproceedings{carion2026sam3,
  author = {Carion, Nicolas and Gustafson, Laura and Hu, Yuan-Ting and
            Debnath, Shoubhik and Hu, Ronghang and Suris Coll-Vinent, Didac and
            Ryali, Chaitanya and Alwala, Kalyan Vasudev and Khedr, Haitham and
            Huang, Andrew and Lei, Jie and Ma, Tengyu and Guo, Baishan and
            Kalla, Arpit and Marks, Markus and Greer, Joseph and Wang, Meng and
            Sun, Peize and R{\"a}dle, Roman and Afouras, Triantafyllos and
            Mavroudi, Effrosyni and Xu, Katherine and Wu, Tsung-Han and Zhou, Yu and
            Momeni, Liliane and Hazra, Rishi and Ding, Shuangrui and Vaze, Sagar and
            Porcher, Francois and Li, Feng and Li, Siyuan and Kamath, Aishwarya and
            Cheng, Ho Kei and Doll{\'a}r, Piotr and Ravi, Nikhila and Saenko, Kate and
            Zhang, Pengchuan and Feichtenhofer, Christoph},
  title = {\href{https://proceedings.iclr.cc/paper_files/paper/2026/hash/e0982cbc81401df3430ee1ff780dc7a2-Abstract-Conference.html}{SAM 3: Segment anything with concepts}},
  booktitle = {International Conference on Learning Representations},
  year = {2026}
}

\clearpage
\onecolumn
\appendices
\renewcommand{\thetable}{S\arabic{table}}
\renewcommand{\theHtable}{supp.\arabic{table}}
\setcounter{table}{0}

\section{Recovery Parameters}
\label{app:implementation-parameters}
The following thresholds and budgets specify DiMuRe's recovery settings
in the reported experiments. Symbols retain their main-text meanings.

\subsection{Program-Execution Grasp Verification}
Table~\ref{tab:supp-grasp} gives the thresholds for the strict inequalities
and evidence checks associated with Eq.~\ref{eq:hold-consistency}.

\begin{table}[!ht]
\centering
\caption{Thresholds for grasp verification during original-program execution.}
\label{tab:supp-grasp}
\begin{tabular}{@{}p{0.17\linewidth}p{0.50\linewidth}p{0.13\linewidth}p{0.12\linewidth}@{}}
\hline
Symbol & Meaning & Value & Unit \\
\hline
$\tau_{\mathrm{move}}$ & Minimum tool displacement for evaluating attachment & 0.025 & m \\
$\tau_{\mathrm{follow}}$ & Maximum accepted object/tool displacement residual & 0.035 & m \\
$g_{\min}$ & Lower admissible gripper opening & 0.003 & m \\
$g_{\max}$ & Upper admissible gripper opening & 0.070 & m \\
\hline
\end{tabular}
\end{table}

\subsection{Recovery Attempts and Physical-Step Budgets}
Table~\ref{tab:supp-budgets} instantiates Eq.~\ref{eq:recovery-budget}.
Attempts are counted before candidate preconditions are checked.
All step quantities refer to native physical control steps.

\begin{table}[!ht]
\centering
\caption{Candidate-attempt limits and physical-step budgets.}
\label{tab:supp-budgets}
\begin{tabular}{@{}p{0.17\linewidth}p{0.50\linewidth}p{0.13\linewidth}p{0.12\linewidth}@{}}
\hline
Symbol & Applicable procedure or invocation & Value & Unit \\
\hline
$A_{\max}$ & Shared candidate-attempt limit across the task & 3 & attempts \\
$b_{\min}$ & Grasp-recovery entry & 250 & steps \\
$b_{\min}$ & Placement-retry entry & 230 & steps \\
$r$ & Grasp recovery and placement retry & 150 & steps \\
$C_{\max}$ & Grasp-recovery procedure & 450 & steps \\
$C_{\max}$ & Placement-retry procedure & 500 & steps \\
$b_{\min}^{\mathrm{vla}}$ & VLA candidate within the ordered grasp route & 420 & steps \\
$C_{\mathrm{req}}^{\mathrm{vla}}$ & Requested VLA cap within the ordered grasp route & 280 & steps \\
$r_{\mathrm{vla}}$ & Bounded VLA tool, including terminal-planner calls & 150 & steps \\
$r$ & Terminal planner's outer procedure & 0 & steps \\
$C_{\max}$ & Terminal planner's outer procedure & $B-n_{\mathrm{start}}$ & steps \\
\hline
\end{tabular}
\end{table}

The outer procedure for terminal replanning can use the budget remaining
at entry. A nested VLA call retains its own 150-step completion reserve
and starts only if at least 120 steps remain after applying the requested
cap, active ceiling, and reserve. VLA requests from the terminal planner
are limited to $[120,260]$ steps. The ordered grasp route requests at most
280 steps.

\subsection{Terminal MLLM-Assisted Replanning}
After normal program exit, the system performs final release checks and
eligible placement retries. Early handoff may instead interrupt the program
after persistent faults, before those retries. The terminal allowance becomes
available before the planner's entry checks. The planner can be invoked at
most once, provided physical execution is still available and the entry
threshold in Table~\ref{tab:supp-planner} is met. Subsequent rounds use the
round-continuation threshold. Its verifier compares successive observations
without using the stored grasp reference in Eq.~\ref{eq:hold-consistency}.

Generic reviews of unverified stop proposals share the cap $K_{\mathrm{review}}$.
Contact-release and skill-availability branches can each request one additional
review; all reviews consume rounds within $K_{\mathrm{plan}}$.

\begin{table}[!ht]
\centering
\caption{Terminal MLLM planner entry thresholds and execution limits.}
\label{tab:supp-planner}
\begin{tabular}{@{}p{0.17\linewidth}p{0.50\linewidth}p{0.13\linewidth}p{0.12\linewidth}@{}}
\hline
Symbol & Meaning & Value & Unit \\
\hline
$b_{\mathrm{entry}}^{\mathrm{plan}}$ & Minimum remaining budget for planner entry & 80 & steps \\
$b_{\mathrm{round}}^{\mathrm{plan}}$ & Minimum remaining budget for beginning a round & 40 & steps \\
$K_{\mathrm{plan}}$ & Maximum planning rounds within the invocation & 32 & rounds \\
$K_{\mathrm{review}}$ & Shared cap on generic reviews of unverified stop proposals & 2 & requests \\
$K_{\mathrm{cmd}}$ & Maximum counted physical-command calls per fragment & 12 & commands \\
$T_{\mathrm{exec}}$ & Wall-clock execution limit per generated fragment & 45 & s \\
\hline
\end{tabular}
\end{table}

The time limit for a generated fragment covers its execution and any calls
made from within it. Automatic observations and model requests outside the
fragment are excluded from this limit but included in end-to-end task time
(Appendix~\ref{sec:supp-metric-definitions}).

\clearpage
\section{Encore Evaluation Protocol and Additional Results}
\label{app:pilot-protocol}

\subsection{Experimental Settings and Configurations}
The completed evaluation contains 4,545 runs with the settings in
Table~\ref{tab:supp-pilot-manifest}. The 801-case subset used in the main text
excludes the 90 cases with 6\,cm receiving-region shifts and retains those
with 12\,cm shifts. This subset was chosen during earlier development and
was not preregistered as a held-out evaluation. Results for both amplitudes
and the full set are reported in
Table~\ref{tab:supp-encore-sensitivity}.

\begin{table}[!ht]
\centering
\caption{Run settings and recovery configurations. Original-program checks
and retries are retained throughout.}
\label{tab:supp-pilot-manifest}
\begin{tabular}{@{}p{0.27\linewidth}p{0.67\linewidth}@{}}
\hline
Setting & Value \\
\hline
Task programs & 30 fixed Encore programs \\
Random seeds & Environment reset: 51, 52, 53; disturbance RNG: $700+\text{seed}$ \\
Evaluation sets & Original/Local/Full: 90 clean and 891 disturbed each; L1-only/L2-only: 801 disturbed each \\
Controlled components & Fixed task programs, perception models, and disturbance settings \\
Disturbance protocol & Trigger rules, perturbation parameters, and validity checks in Appendix~\ref{app:benchmark-evidence} \\
Recovery configurations & Original: no added recovery; L1-only: program reuse; L2-only: geometry/VLA skills; Local: L1+L2; Full: Local plus terminal replanning \\
Configuration differences & L2-only disables mixed program-segment placement retry; Full enables early handoff \\
Physical-step allowance & Original/Local/L1-only/L2-only: 1,000; Full: 1,000 native + 400 terminal \\
Execution conditions & Two GPUs, one episode per GPU; shared SAM~3~\cite{carion2026sam3} perception and GraspVLA~\cite{deng2025graspvla} services \\
Terminal model & DeepSeek V4.1 Flash \\
\hline
\end{tabular}
\end{table}

Full raises the cumulative step ceiling from 1,000 to 1,400 without resetting
the counter, so unused native steps remain available. Because Full also
changes the budget and handoff timing, its gains over Local cannot be
attributed to L3 alone. Disabling the mixed placement retry in L2-only
also prevents fully isolating the effect of L1.

\subsection{Metric Definitions}
\label{sec:supp-metric-definitions}
Table~\ref{tab:supp-metric-definitions} defines each metric and the episodes
included in its denominator. Rates are shown as percentages. A zero denominator
is reported as N/A, and unmeasured values are not replaced by zero.

The task interval begins with the first online observation or program operation,
whichever occurs first. It ends at the first native success or, for a failed
run, the first recorded native termination. If a failed run has no termination
record, the interval ends when online execution ends. Program exit or handoff
to the planner does not by itself end the interval. Task time includes sensing,
preprocessing, service waits, inference, execution, and retries. It excludes
reset, loading, warm-up, and offline program construction. Physical steps count
executed native control steps within the same interval, including recovery.
Simulator substeps and high-level calls are not counted as control steps.

\begin{table}[!ht]
\centering
\caption{Metric definitions and statistical populations. All task-success
measures use native complete-task success; local grasp verification is separate.}
\label{tab:supp-metric-definitions}
\small
\renewcommand{\arraystretch}{1.15}
\begin{tabular}{@{}p{0.25\linewidth}p{0.71\linewidth}@{}}
\hline
Metric & Definition and denominator \\
\hline
SR and $\Delta$SR & Native task successes divided by all scheduled episodes,
including unreached or ineffective disturbances. $\Delta$SR is the difference
from Original in percentage points, computed from unrounded rates. \\
Recovery and retention & Successful outcomes among Original failures and
among Original successes, respectively, paired by scheduled case. \\
Mean task time and steps & Arithmetic means over all scheduled episodes,
including failures, or over each configuration's successful subset as labeled.
Successful subsets can differ between configurations. \\
Recorded recovery time & Sum of logged stage intervals within the task
interval, averaged over episodes entering that stage before the endpoint,
including failed episodes. Local and terminal stages are measured separately. \\
Anomaly-response coverage & Auditable first-fault windows containing a
qualifying alarm or target adaptation, divided by all auditable windows;
repeated responses count once. \\
Task SR after response & Native task successes strictly after the first
qualifying response, divided by responding windows. \\
Diagnostic accuracy & Correct class-and-role diagnoses in the final eligible
formal output, divided by all auditable first-fault windows. \\
Task SR after correct diagnosis & Subsequent native task successes divided by
correctly diagnosed windows, including windows in which recovery is not executed. \\
Terminal-model use & Submitted L3 requests and recorded tokens averaged over
all scheduled episodes, including episodes without planner entry and requests
whose responses are rejected. Cached input is already included in input tokens. \\
\hline
\end{tabular}
\end{table}

Recorded recovery time includes sensing, reasoning, waits, actions, and
verification within each stage. These costs are already included in task time.
The logs do not separately time routine monitoring or inline adaptations
without recorded start and end times, so this measure does not capture all
diagnosis and recovery overhead. Each stage's mean uses its own episode set,
and an episode may enter more than one stage.

Time and step records are complete for the 801-case subset. Successful-case
means use 470, 590, 525, 622, and 656 episodes for Original, L1-only, L2-only,
Local, and Full, respectively. Configurations shared services, and their
runs were not fully interleaved. Wall-clock comparisons are therefore descriptive.

\subsection{Full Evaluation Set and Disturbance Verification}
The task-success denominator includes all scheduled cases, even when a
trigger is not reached, a disturbance has no effect, or execution ends because
of budget exhaustion or a safety stop. Table~\ref{tab:supp-encore-sensitivity}
also reports the 6\,cm cases, on
which Local and Full complete fewer tasks than Original.

\begin{table}[!ht]
\centering
\caption{Native task success on the full disturbed development set,
the main-text subset, and each receiving-region shift amplitude.
Entries are successful episodes / scheduled episodes. Cases remain in
the denominator whether or not the intervention has a verified effect.}
\label{tab:supp-encore-sensitivity}
\small
\setlength{\tabcolsep}{8pt}
\begin{tabular}{@{}lrrrr@{}}
\hline
Evaluation set & $N$ & Original & Local & Full \\
\hline
Main-text subset (without 6\,cm) & 801 & 470/801 & 622/801 & 656/801 \\
Full disturbed set & 891 & 549/891 & 689/891 & 725/891 \\
Receiving-region shift, 6\,cm & 90 & 79/90 & 67/90 & 69/90 \\
Receiving-region shift, 12\,cm & 90 & 49/90 & 77/90 & 77/90 \\
\hline
\end{tabular}
\end{table}

Table~\ref{tab:supp-encore-audit} distinguishes whether an intervention was
\emph{applied} from whether its physical effect was verified (\emph{realized}).
A missing effect check remains unknown. The official goal predicate is also
checked after each run. This final-time check is separate from the native
evaluator's complete-task success signal.

\begin{table}[!ht]
\centering
\caption{Disturbance application, effect verification, and final-time checks
on 801 scheduled cases per configuration. Application records are complete.
Effect checks are either positive or missing; missing checks remain unknown.}
\label{tab:supp-encore-audit}
\small
\setlength{\tabcolsep}{8pt}
\begin{tabular}{@{}lrrrr@{}}
\hline
Configuration & Applied
 & \shortstack{Effect\\verified}
 & \shortstack{Effect check\\missing}
 & \shortstack{Final check\\passed} \\
\hline
Original & 631 & 105 & 696 & 470 \\
L1-only & 632 & 105 & 696 & 590 \\
L2-only & 632 & 105 & 696 & 523 \\
Local DiMuRe & 632 & 105 & 696 & 620 \\
Full DiMuRe & 632 & 105 & 696 & 654 \\
\hline
\end{tabular}
\end{table}

Across all 4,545 runs, eight episodes achieve native success but fail the
final-time check: three Local runs, their three matched Full runs, and two
L2-only runs. No run passes the final-time check without achieving native
success. Task SR uses the native evaluator's complete-task success signal.
The available records do not explain these discrepancies.

\subsection{Paired and Per-Type Outcomes}
Table~\ref{tab:supp-paired-outcomes} measures how often each configuration
recovers a case that Original fails or retains a case that Original completes.
The denominators are Original's failures and successes, respectively.

\begin{table}[!ht]
\centering
\caption{Paired native outcomes on the 801-case disturbed subset.
Counts and percentages use Original's failures (left) or successes (right)
as the denominator.}
\label{tab:supp-paired-outcomes}
\small
\setlength{\tabcolsep}{10pt}
\begin{tabular}{@{}lrr@{}}
\hline
Configuration & \shortstack{Original failures\\recovered (\%)}
 & \shortstack{Original successes\\retained (\%)} \\
\hline
Full DiMuRe & 194/331 (58.61) & 462/470 (98.30) \\
Local DiMuRe (no L3) & 159/331 (48.04) & 463/470 (98.51) \\
L1-only & 125/331 (37.76) & 465/470 (98.94) \\
L2-only & 69/331 (20.85) & 456/470 (97.02) \\
\hline
\end{tabular}
\end{table}

Relative to Local, Full succeeds on 35 additional cases and fails on one
case that Local completes. In that lost case, Full invokes early handoff
while the matched Local run continues to success. Comparing the single-level
configurations, 106 cases succeed only with L1-only and 41 only with L2-only.
Local succeeds on three cases where both single-level configurations fail
and fails on two where both succeed. Runs are paired by scheduled case,
but their subsequent states can differ. Table~\ref{tab:supp-single-level-types}
reports outcomes by disturbance type.

\begin{table}[!ht]
\centering
\caption{Native task successes by scheduled disturbance type on the
801-case development subset. Each entry is a success count with the row
denominator $N$. Full has a larger allowance and uses early handoff.
Scheduled cases remain in the denominator even when no disturbance is applied.}
\label{tab:supp-single-level-types}
\small
\setlength{\tabcolsep}{6pt}
\begin{tabular}{@{}lrrrrrr@{}}
\hline
Disturbance & $N$ & Original & L1-only & L2-only & Local & Full \\
\hline
Arm retraction & 90 & 83 & 83 & 84 & 84 & 84 \\
Grasp slip & 90 & 38 & 72 & 49 & 70 & 76 \\
Object bump & 90 & 90 & 90 & 90 & 90 & 90 \\
Object pose change & 81 & 26 & 26 & 47 & 47 & 47 \\
Receiving-region shift (12\,cm) & 90 & 49 & 77 & 49 & 77 & 77 \\
Object retrieval & 90 & 24 & 58 & 29 & 58 & 69 \\
Slow object drift & 90 & 85 & 85 & 83 & 85 & 88 \\
Object tipover & 90 & 43 & 44 & 56 & 55 & 59 \\
Object XY displacement & 90 & 32 & 55 & 38 & 56 & 66 \\
\hline
Total & 801 & 470 & 590 & 525 & 622 & 656 \\
\hline
\end{tabular}
\end{table}

\subsection{Timing Audit of Recorded Recovery Procedures}
\label{sec:supp-recovery-timing}
The timing analysis uses all 801 logs per configuration, regardless of fault-label
availability. All procedure entry and exit times are recorded in order.
Within the task interval defined in Appendix~\ref{sec:supp-metric-definitions},
271 episodes enter local recovery in each of Local and Full, and 180 Full
episodes enter terminal recovery. Local and terminal intervals do not overlap
in time, although an episode can enter both stages.

\subsection{Retrospective Diagnostic Evaluation}
\paragraph{Independent labels and coverage}
In each of Local and Full, 161 of 801 episodes (20.10\%) contain a first-fault
window that can be labeled independently. Labels were assigned from
ground-truth (GT) states and video from the same trial, without access to
diagnostic outputs, disturbance names, or final task outcomes. Each label
identifies the earliest fault supported by the evidence and the role of the
affected object. Naturally occurring faults are included. Labels were fixed
before comparison with diagnostic outputs. The other 640 episodes per
configuration lack sufficient evidence about the fault or its timing, contain
ambiguous concurrent faults, or undergo recovery before the first fault can
be established. These episodes remain unlabeled.

A fault window starts at fault onset and ends when recovery first changes
the faulty condition, the condition disappears, or the task terminates.
Taking a new observation alone does not end the window. GT steps and diagnostic timestamps
are aligned using the recorded start of online execution, native control-step
records, and video frame indices. If onset is uncertain, an output is accepted
only if it occurs after the latest possible onset or adjacent step records
establish that it follows the fault. Unresolved temporal orderings remain unscorable.

\paragraph{Response and diagnostic criteria}
A response counts if it occurs within the fault window and either raises
an execution-anomaly alarm or adjusts a grasp or placement target using new
observations. This includes an adjustment that ends the window.
Insufficient-evidence outputs, positive checks, budget-status messages,
recovery entry without a qualifying response, and alarms after task success
do not count.

Diagnostic accuracy uses the last eligible formal output before the window
ends. That output must match both the labeled fault class and the object role.
Missing, incorrect, or unresolved outputs count as incorrect, even if an
earlier output was correct. All correctly diagnosed windows enter the
denominator for subsequent task success, including those in which recovery
is not admitted or executed.

\begin{table}[!ht]
\centering
\caption{Anomaly responses, formal diagnoses, and subsequent native task
success on 161 auditable windows per configuration. Entries show numerator /
denominator. The shared columns have identical observed counts for Local and
Full; success after response and after correct diagnosis use different populations.}
\label{tab:supp-diagnostic-audit}
\small
\setlength{\tabcolsep}{5pt}
\begin{tabular}{@{}lrrrrr@{}}
\hline
Fault family & \shortstack{Response\\(both)}
 & \shortstack{Correct diagnosis\\(both)}
 & \multicolumn{2}{c}{Task success after response}
 & \shortstack{Task success after\\correct diagnosis\\(both)} \\
\cline{4-5}
 & & & Local & Full & \\
\hline
Grasp not established & 31/34 & 15/34 & 23/31 & 24/31 & 15/15 \\
Holding lost & 30/34 & 12/34 & 15/30 & 21/30 & 11/12 \\
Receiving region moved & 31/36 & 30/36 & 28/31 & 28/31 & 28/30 \\
Target state changed & 34/57 & 0/57 & 28/34 & 28/34 & N/A \\
\hline
Total & 126/161 & 57/161 & 94/126 & 101/126 & 54/57 \\
\hline
\end{tabular}
\end{table}

Both configurations have 78.26\% response coverage and 35.40\% formal
diagnostic accuracy (Table~\ref{tab:supp-diagnostic-audit}). Task success
after a correct diagnosis is 94.74\%. A response may address a later problem
in the window without correctly identifying the first fault. In the target-state
category, no formal diagnosis matches the label (0/57), although target
adjustments can still count as responses.

These conditional success rates do not establish a causal effect of recovery.
Labels use object roles because diagnostic logs do not identify individual
object instances. Normal states were not exhaustively labeled, so these
rates cannot be generalized to all 801 episodes or used to estimate false
alarms. The logs also do not consistently record when diagnostic processing
begins, and some fault onset times remain uncertain. They therefore do not
support complete diagnostic-latency measurements or reliably aligned
onset-to-alarm delays.

\subsection{Terminal MLLM Use}
Full invoked the terminal planner in 180 of 801 disturbed episodes and
submitted 1,547 requests, including requests whose responses were rejected.
Original and Local had terminal replanning disabled and made no L3 requests.
Table~\ref{tab:supp-model-use} covers terminal-model requests only;
usage by perception, VLA, and diagnosis-only calls was not fully recorded.

\begin{table}[!ht]
\centering
\caption{Full DiMuRe terminal-model use, averaged over all 801 scheduled
episodes, including those without planner entry. Cached input is included
in input tokens. Values are rounded from recorded request usage.}
\label{tab:supp-model-use}
\small
\setlength{\tabcolsep}{9pt}
\begin{tabular}{@{}lrrrr@{}}
\hline
Configuration & \shortstack{L3\\requests}
 & \shortstack{Input\\tokens}
 & \shortstack{Output\\tokens}
 & \shortstack{Cached input\\tokens} \\
\hline
Full DiMuRe & 1.93 & 204,421 & 570 & 30,825 \\
\hline
\end{tabular}
\end{table}

\section{LIBERO-Disturb Protocols and VLA Evidence}
\label{app:benchmark-evidence}

\subsection{Disturbance Definitions and Encore Settings}
All nine disturbance types in Fig.~\ref{fig:libero-disturb-nine} are represented
in the Encore evaluation. Table~\ref{tab:supp-disturbance-coverage} lists their
triggers and intended interventions in the same order as the figure.
Object pose change is configured for 27 programs (81 cases). Each other
type has 90 cases per amplitude. Receiving-region shifts of 6 and 12\,cm
are two settings of one type, giving 891 disturbed cases in total.
The 801-case main-text subset retains only the 12\,cm setting.

\begin{table}[!ht]
\centering
\caption{Nine disturbance types in the evaluated Encore protocol.
Steps are native physical control steps. The settings describe intended
interventions, not guaranteed physical effects or task failures.}
\label{tab:supp-disturbance-coverage}
\small
\begin{tabular}{@{}p{0.18\linewidth}p{0.33\linewidth}p{0.43\linewidth}@{}}
\hline
Disturbance & Trigger or target selection & Nominal intervention \\
\hline
Object XY displacement & Step 30; declared task target & Random horizontal shift of 6--10\,cm; bounded by workspace, separation, and penetration checks \\
Object bump & Step 40; random free object excluding the nearest to the tool & Horizontal velocity increment of 0.3--0.6\,m/s in a random direction \\
Slow object drift & Step 20; object nearest to the tool & Horizontal translation of 1\,mm per step for 60 steps in a random direction \\
Object retrieval & Poll every 5 steps from step 80; held object displaced more than 8\,cm in XY & Reset the object's position to its initial position and zero its velocity \\
Object pose change & Poll every 10 steps from step 20 to 520; eligible resting target & Apply the predefined orientation assigned to the case, subject to clearance and settling checks \\
Object tipover & Step 40; object nearest to the tool & Rotate by $90^\circ$ about a random horizontal axis and raise by 3\,cm \\
Grasp slip & Poll every 5 steps during steps 60--500; held-object eligibility & Lower the object by 3--5\,cm and shift each horizontal coordinate by at most 2\,cm \\
Arm retraction & Stochastic trigger during steps 30--149 & Add independent joint-position offsets in $[-0.15,0.15]$\,rad to seven arm joints \\
Receiving-region shift & Poll every 10 steps from step 20; held object transported at least 4\,cm in XY, tool--receiver XY distance at least 15\,cm & Move the declared receiver along world $+X$ by 6 or 12\,cm over 18 steps \\
\hline
\end{tabular}
\end{table}

The nearest object is selected by horizontal distance to the tool and may
differ from the task target. The injector treats an object as held only if
its height has increased by more than 5\,cm and the mean finger-joint position
is below 4\,cm. This trigger rule differs from the recovery system's grasp
verification. For arm retraction, a new trigger step is sampled at each step.
The intervention occurs only when the sampled index equals the current one.
Each episode contains at most one intervention event, which may span several steps.

Proposed XY displacements are clipped to the horizontal workspace
$[-0.35,0.35]$\,m per axis. A proposal is accepted only if the resulting shift
is 5--12\,cm, the horizontal distance to the tool is at least 10\,cm, and
the horizontal distance to other objects is at least 7\,cm. If penetration
into a non-supporting object exceeds 5\,mm, the proposal is rejected and
the previous state restored. Pose changes use the orientation fixed for
each case, require a tool distance of 15--25\,cm, and check clearance at
nine points along the rotation. A 1\,s settling test then holds the robot
fixed and restores the simulation state after the test, before applying the
accepted intervention. Drift and receiver shifts are also clipped to the
workspace. These checks use simulator state within the injector. That state
is not supplied to online diagnosis or recovery. Application and effect
verification are reported separately in Table~\ref{tab:supp-encore-audit}.

\subsection{Native VLA Execution Protocol}
The separate $\pi_{0.5}$ evaluation contains 120 episodes: the same 40 tasks
(30 short and 10 long) under clean, empty-hook, and receiving-region-shift
conditions. The empty-hook condition enables the disturbance interface without
applying a perturbation. This evaluation covers one disturbance type using a
policy-specific injector. It does not evaluate all nine types or apply DiMuRe
recovery to the VLA policy. The adapter preserves the policy's original action
queue and control clock and adds no settling steps, queue clearing, replanning,
or recovery assistance. Fixed settings are given in
Table~\ref{tab:supp-vla-protocol}.

\begin{table}[!ht]
\centering
\caption{Fixed settings of the native $\pi_{0.5}$ evaluation.}
\label{tab:supp-vla-protocol}
\begin{tabular}{@{}p{0.34\linewidth}p{0.60\linewidth}@{}}
\hline
Setting & Value or rule \\
\hline
Initial state & Official state index 0 for each task \\
Random seeds & Environment 7; policy 7; disturbance 51 \\
Native physical-step limits & Spatial: 220; Object: 280; Goal: 300; Long: 520 \\
Initialization and action queue & 10 native wait steps; 5-action chunks \\
Trigger polling & Begin after 20 policy steps; retry after 10 steps if rejected \\
Transported-object eligibility & Lift $\geq 2.5$\,cm; tool-to-object distance $\leq 13$\,cm; closing command; bilateral contact for 3 consecutive steps \\
Transport condition & Target XY displacement from its initial position $\geq 4$\,cm \\
Relocation eligibility & Empty, unambiguously identified free-joint receptacle; tool distance $\geq 15$\,cm; valid full movement path \\
Target and event count & First in/on placement relation specified by the task; at most one applied event per episode \\
Nominal relocation & World $+X$: 12\,cm over 0.9\,s; Long task 0: 9\,cm over 0.9\,s \\
\hline
\end{tabular}
\end{table}

\subsection{VLA Outcomes and Disturbance Verification}
Table~\ref{tab:supp-vla-results} includes all 40 scheduled tasks in each
condition, including cases with inapplicable or unreached triggers.
An episode counts as successful only when a completed native control step
returns the task-success signal. Satisfying a goal predicate in the final
state, including after a safety stop, does not replace this criterion.

\begin{table}[!ht]
\centering
\caption{Complete native-task outcomes in the reported $\pi_{0.5}$ subset.
SR is the complete-task success percentage. Applied counts indicate that
an intervention began, without implying that its intended effect was verified.
Safety-stopped runs count as unsuccessful.}
\label{tab:supp-vla-results}
\begin{tabular}{lrrrrr}
\hline
Condition & Runs & Successes & SR (\%) & Applied & Safety stops \\
\hline
Clean & 40 & 39 & 97.5 & 0 & 0 \\
Empty hook & 40 & 39 & 97.5 & 0 & 0 \\
Receptacle relocation & 40 & 36 & 90.0 & 22 & 1 \\
\hline
\end{tabular}
\end{table}

A relocation is applied when at least one movement increment is executed.
Of the 22 applied events, 21 complete the prescribed motion with endpoint
error at most 1\,cm and one is partial. The 18 unapplied cases comprise
13 inapplicable cases and five unreached triggers. Application, completed
shift, and native task success are recorded separately.

Each episode is linked to its own video, native trajectory, and intervention
log. The audit compares clean and empty-hook trajectories and checks trajectory
agreement before injection. Calibration replays and preliminary validation
runs are excluded from task-success denominators. Encore and VLA results
are reported separately because their triggers, allowances, and execution
protocols differ.

\end{document}